\documentclass{llncs}

\usepackage{graphicx}
\usepackage[T1]{fontenc}
\usepackage[latin1]{inputenc}

\begin{document}

\title{An optimal linear separator for the Sonar Signals Classification task}

\author{Juan-Manuel Torres-Moreno\inst{1,}\inst{2} \and Mirta B. Gordon\inst{3}}

\authorrunning{Torres-Moreno and Gordon}

\institute{Laboratoire Informatique d'Avignon\\
Université d'Avignon et des Pays de Vaucluse\\
BP 1228 84911 Avignon Cedex 09, France\\
\and
 \'Ecole Polytechnique de Montréal\\
 CP 6079 Succ. centre-Ville, H3C 3A7 Montréal, Québec Canada\\
 \and
 TIMC Domaine de La Merci, \\
38706 La Tronche Cédex France\\
\email{juan-manuel.torres@univ-avignon.fr}
}


\date{july 1996}

\maketitle

\begin{abstract} 

\noindent The problem of classifying sonar signals from rocks and mines first studied by Gorman and Sejnowski has become a benchmark against which many learning algorithms have been tested. We discovered that both the training set and the test set of this benchmark are linearly separable, although with different hyperplanes. Moreover, the complete set of learning and test patterns together, is also linearly separable. We give the weights that separate these sets, which may be used to compare results found by other algorithms.

\end{abstract}


\section{Introduction}

It has become a current practice to test the performance of learning algorithms on \textsl{realistic} benchmark problems. 
The underlying difficulty of such tests is that in general these problems are not well caracterized, making it thus impossible to decide whether a better solution that the one already found exists. 

The Sonar signals classification benchmark, introduced by Gorman et al. \cite{Sonar} is widely used to test machine learning algorithms. 
In this problem the classifier has to discriminate if a given sonar return was produced by a metal cylinder or by a cylindrically shaped rock in the same environment. 
The benchmark contains 208 preprocessed sonar spectra, defined by $N=60$ real values, with their corresponding class. 
Among these, $P=104$ patterns are usually used to determine the classifier parameters through a procedure called learning. 
Then, the classifier is used to class the $G=104$ remaining patterns and the fraction of misclassified patterns is used to estimate the generalization error produced by the learning algorithm. 
We applied Monoplane, a neural incremental learning algorithm, to this benchmark. 
In this algorithm, the hidden units are included one after the other until the number of training errors vanishes. 
Each hidden unit is a simple binary perceptron, trained with the learning algorithm Minimerror \cite{Minimerror}.

\section{The Incremental Learning Algorithm}

\subsection {Definitions}

Consider a training set of $P$ input-output pairs $\{\vec{\xi}^{\mu}, \tau ^{\mu}$\}, where $\mu=1,2,\cdots,P$. 
The inputs $\vec{\xi}^{\mu}=(1, \xi_1^{\mu}, \xi_2^{\mu}, \cdots, \xi_N^{\mu})$ may be binary or real valued $N+1$ dimensional vectors. 
The first component $\xi_0^{\mu}=1$, the same for all the patterns, allows to process the bias as a supplementary weight. 
The outputs are binary, $\tau^{\mu}= \pm 1$. 
The Neural Network built by the learning algorithm has a single hidden layer of $H$ binary neurons connected to the $N+1$ input units, and one output neuron connected to the hidden units. 
During training, the number of hidden neurons grows until the number of training errors vanishes. 
After learning, the hidden units $1 \leq h \leq H$ have synaptic weights $\vec w_h=(w_{h0}, w_{h1} \cdots w_{hN})$, $w_{h0}$ being the bias of unit $h$. 
The output neuron has weights $\vec W = (W_0, W_1 \cdots W_h)$ where $W_0$ is the bias.

Given an input pattern $\vec{\xi}$ that the network has to classify, the state $\sigma_h$ of hidden neuron $h$ is given by:

\begin{equation} 
\label{eq:update_h} \sigma_h = sign \left(\sum_{i=0}^N w_{hi} \xi_i \right) ; \;\;\;\;\;\;\;\; h=1,\cdots,H 
\end{equation}

\noindent The network's output $\zeta$ is given by:

\begin{equation} \label{eq:update_output} \zeta = sign
\left(\sum_{h=0}^H W_h \sigma_h \right) \end{equation}

\noindent with $\sigma_0=1$ for all the patterns.

\subsection{Monoplane algorithm}

The Monoplane algorithm \cite{Monoplane,thesis_torres} constructs a hidden layer in which each appended hidden unit tries to correct the training errors of the previous hidden unit. 
\noindent In fact, the construction of hidden layer is similar to the first layer construction of the parity machine \cite{Offset,Tiling-Parity}. But, instead of introducing a second hidden layer implementing the parity, our algorithm goes on adding hidden units (if it is necessary). 
In the case of binary inputs it was proven \cite{Offset} that a solution exists with at most $P$ hidden neurons. 
A solution for real valued inputs also exists \cite{Convergence}, the upper bound to the number of hidden units being $P-1$. 
The proof that a solution with a finite number of units also exists, is found in \cite{Convergence}. 
Thus, the algorithm Monoplane converges to a finite size network. 
Clearly, the upper bounds are not tight, and in practice the algorithm constructs very small Neural Networks \cite{thesis_torres}.

The final number $H$ of hidden units depends on the performance of the learning algorithm used to train the individual binary perceptrons. The best solution should endow the perceptron with the lowest generalization error if the training set is LS, and should minimize the number of errors otherwise. Most incremental strategies use the Pocket algorithm \cite{Pocket}. 
It has no natural stopping condition, which is left to the user's patience. 
None of the proposed alternative algorithms as \cite{Thermal_perceptron} are guaranteed to find the best solution to the problem of learning. 
The success of our incremental algorithm relies on the use of Minimerror to train the individual units. 
Minimerror is based on the minimization of a cost function $E$ which depends on the weights $\vec w$ through the stabilities of the patterns of the training set. 
If the input vector is $\vec{\xi}^{\mu}$ and $\tau^{\mu}$ the corresponding target, then the {\sl stability} $\gamma^{\mu}$ of pattern $\mu$ is a continuous and derivable function of the weights, given by~:

\begin{equation} \label{eq:stability} \gamma^{\mu} = \tau^{\mu}
\frac{\vec w \cdot \vec{\xi}^{\mu}}{ \parallel \vec{w} \parallel }
\end{equation}

\noindent where $\parallel \vec{w} \parallel = \sqrt{\vec w \cdot \vec w}$. 
The stability measures the distance of the pattern to the separating hyperplane normal to $ \vec{w}$~; it has a positive sign if the pattern is well classified, negative otherwise. 
The cost function $E$ is given by:

\begin{equation} \label{eq:cost} E=\frac{1}{2} \sum_{\mu=1}^P \left[
1- \tanh \frac{\gamma^{\mu}}{2T}\right] \end{equation}

\noindent The contribution to $E$ of patterns with large negative stabilities is 1, i.e. they are counted as 1 error, whereas the contribution of patterns with large positive stabilities is vanishingly small. 
Patterns within a window of width $\approx 2T$ centered on the hyperplane contribute to the cost function even if they have positive stability, proportionally to $1-\gamma/T$. 
It may be shown that $E$ may be interpreted as a noisy measure, at temperature $T$, of the number of training errors \cite{GorPerBer}. 
The properties of its global minimum, studied theoretically with methods of statistical mechanics \cite{GorGre}, have been confirmed by numerical simulations \cite{Minimerror,RaffGor}. 
In particular, the minimum of $E$ in the limit $T \rightarrow 0$ corresponds to the weights that minimize the number of training errors. 
If the training set is LS, the weights that separate the training set are not unique. 
It was shown that there is an optimal learning temperature such that the minimum of the cost function endows the perceptron with a generalization error numerically indistinguishable from the optimal (bayesian) value.

The algorithm Minimerror minimizes the cost $E$ through a gradient descent, combined with a slow decrease of the temperature $T$ equivalent to a deterministic annealing \cite{Minimerror,RaffGor}, and determines automatically the optimal temperature at which it has to stop.

\section{The Sonar Benchmark} 

The set of exemples contains 111 patterns obtained by bouncing sonar signals off a metal cylinder at various angles, and 97 patterns obtained from rocks under 
similar conditions. Each pattern is a set of 60 numbers in the range [0,1]. 
Each number represents the energy within a particular frequency band, 
integrated over a certain period of time. The label associated with each 
pattern contains the number $\tau=+1$ if the signal correspond to a rock and 
$\tau=-1$ if it is a mine (metal cylinder).  

\begin{table}
\begin{center}
\begin{tabular}{|c|c|c|c|c|c|}
\hline
  SET               & $N$ & $P$ & $G$ & $\tau=+1$ & $\tau=-1$ \\
\hline
$Train$             & 60  & 104 & 104 &  55       & 49 \\
\hline
$Test$              & 60  & 104 & 104 &  42       & 62 \\
\hline
$Sonar(Test+Train)$ & 60  & 208 & 0   &  97       & 111 \\
\hline
\end{tabular}
\caption{Number of patterns and distribution of classes}
\end{center}
\end{table}

We have numbered each patterns with a label absolute $\mu$. Of this way, the set $Train$ has $\mu $ of 1 to 104 and the set $ Test $, $ \mu $ of 105 to 208. 
This identification allows to analyze each pattern of way infividual. 
The used procedure was the following: learning the $P$ patterns of $Train$ set and measure the error of generalization on the $G$ patterns of $Test$ set. 
Later, learn on the $P$ patterns of $Test$ set and measure generalization over $G$ patterns of $Train$ set. Finally take the $Sonar$ set ($Train+ Test$) 
 and try to learn it. In this last case, of course there is not possibility of measuring generalizacion.
We have carried out a pre-processing of the vector $\vec {\xi}$ of the learning set, through the following normalization:

\begin{equation}
  \xi_i^{\mu} \leftarrow    \frac{\xi_i^{\mu} - \langle \xi_i \rangle}{\sigma_i}
\end{equation}

\begin{equation}
  \langle\xi_i\rangle = \frac { \sum_{\mu=1}^P \xi_i^\mu} {P}
\end{equation}
  
\begin{equation}
  \sigma_i = \frac { \sum_{\mu=1}^P {(\xi_i^\mu - \langle\xi_i\rangle)}^2 } {P}
\end{equation}

Some authors \cite{CC/limited,Roy} have reported that the learning set $Train$ is linearly separable. But most of people, report results obtained through the backpropagation algorithm (or their variants), and they find too complex nets: with a number excessive of parameters \cite{Verma}(weigths and units).

We found \cite{Monoplane} that we needed two hidden units to learn without errors the training set $Train$, but the generalization error was lower with only one unit (a simple perceptron) than with two hidden units. This result is usually considered as overfitting, a not well defined category used to describe this kind of behaviour.

A finer tuning of the parameters of Minimerror showed however that this benchmark is linearly separable. 
In fact, the $Train$ set, the $Test$ set, and {\it both} sets together (i.e. the $P+G=208$ in $Sonar$ set) patterns are linearly separable.
In Tables \ref{table:Weights_train}, \ref{table:Weights_test} and \ref{table:Weights_sonar}, we give the values for the weights of the perceptron that separate each of the three sets $Train$, $Test$ and $Sonar$.

\begin{table}
\begin{tabular}{|cl|}
\hline
W$_{Train} =$  \{ & \\
&   -0.0692, -1.5031, -1.9481, -0.2835, -1.0162, -0.2870, -0.5139 -0.3040, \\
&  2.3106, -0.4349, -0.6610, -1.0995, -1.2447, -1.3281,  -0.7392,  0.6469,\\
&  1.7862,  1.2227, -0.0513, -0.6431, -0.8745, -0.8290, -0.8084, -0.6578,\\
& -1.0453, -1.2332, -0.9860, -1.0617, -1.0097, -1.3597, -0.5245,  1.6822,\\
&  0.6588, -0.1056, -0.0794, 0.2998,  1.2290,  0.6709, -0.3025,  0.2681,\\
&  1.2375,  0.2485, 0.1098,  0.1693, -0.5717, -1.2458, -0.7116, -0.1323,\\
& -1.3481, -2.6467,  1.0464, -0.7163, -0.8324, -0.4364, -1.1849,  1.3439 \\ 
&  0.4299,  1.0813, -0.9662, -0.3129,  0.0015 \\
& \} \\
\hline
\end{tabular}
\caption{Weights of Minimerror trained perceptron for $Train$ test}
\label{table:Weights_train}
\end{table}

\begin{table}
\begin{tabular}{|cl|}
\hline
$W_{Test} =$ \{ & \\
&   -0.4035, -0.9738,  1.0107,  0.9301, -0.8997, -0.5649,  0.9318,  1.5102,\\
&   0.0477, -1.6914, -1.3137, -2.0763, -2.0756, -0.5307,  0.8317,  1.4271, \\
&   0.6112,  0.5119,  0.2081, -0.8285, -1.4488, -1.4337, -1.1908, -1.0213,  \\
&  -0.3653,  0.2701,  0.2465, -0.2028, -0.3975, -0.2049,  0.1843,  1.2486, \\
&   0.3270,  0.2806,  0.4427,  0.7089,  1.5015,  1.5818,  0.2483, -0.6511, \\
&   0.6822,  0.4056, -0.4476, -1.4451, -2.1873, -1.5600, -1.0694, -0.6042, \\
&  -0.5170, -0.1298,  1.0330, -1.3454, -1.6560,  0.1098, -0.1249, -0.0331, \\
&  -0.1748,  0.2088, -0.7949, -1.7304,  0.1419 \\
& \}\\
\hline
\end{tabular}
\caption{Weights of Minimerror trained  perceptron $Test$ set}
\label{table:Weights_test}
\end{table}

\begin{table}
\begin{tabular}{|cl|}
\hline
$W_{Sonar} =$ \{ & \\
&   -0.0290, -0.7499, -0.0626,  0.5991, -0.1493,  0.2057, -0.3432,  0.5235, \\
&    0.2407, -0.2480, -0.2069,  0.4854, -2.0100,  0.7256,  0.0868, -0.6282, \\
&    1.0116,  0.8600, -0.8842, -0.1056, -1.4125,  1.7174, -2.1101,  0.3378, \\
&   -0.8198, -0.1804,  1.4065, -2.2840,  1.4039, -0.1153, -2.6714,  3.3527, \\
&   -1.0352, -1.1619,  1.4134, -0.6482,  0.3479,  0.9895, -0.3477, -0.5707, \\
&    1.2758, -0.6628,  0.6288, -0.7920, -0.0850, -0.1348, -0.7794,  0.2451, \\
&   -0.8392, -0.5660,  1.4128, -0.4471, -0.5439, -0.2079, -0.1840, -0.0060, \\
&    0.2276, -0.0158, -0.2637, -0.1579,  0.1238 \\
& \}\\
\hline
\end{tabular}
\caption{Weights of Minimerror trained perceptron that separates the Full $Sonar$ ($Train$ + $Test$) dataset}
\label{table:Weights_sonar}
\end{table}

We have calculated in table \ref{table:cos} the cosine of the angle $\alpha$ between the vector $W_{Sonar}$ that separates the whole set, and the $W_{Train}$, $W_{Test}$ 
vectors, that separate $Test$ and $Train$ set respectively. 
Also the cosine between  $W_{Train}$ and $W_{Test}$ is calculated, following equation (\ref{eq:cosinus}).

\begin{equation}
cos(\alpha) = \frac {\vec {{W}_{a}} \cdot \vec {W}_b } { (N+1)^2 } 
\label{eq:cosinus}
\end{equation} 

\begin{table}
\begin{center}
\begin{tabular}{|c|c|c|c|}
\hline
 &      $(W_{Sonar},W_{Train})$ & $(W_{Sonar},W_{Test})$ & $(W_{Train},W_{Test})$ \\
\hline
$cos(\alpha)$ & 0.51615  & 0.34238   &  0.4   \\ 
\hline
\end{tabular}
\caption{Cosine}
\label{table:cos}
\end{center}
\end{table}

In Table \ref{tab:Stabilities_tt} at right, we give the distances of each pattern bad classified by $W_{Train}$ to the hyperplane separator $W_{Sonar}$.
$\epsilon_{g}$ = 19.2 (15 F+   5  F-). At left, we give the distances of each pattern bad classified by $W_{Test}$ to the hyperplane separator $W_{Sonar}$.
$\epsilon_{g}$ = 23.1 (5 F+   19  F-)

\begin{table}
\begin{minipage}[t]{.4\linewidth}
\begin{tabular}{|c|c|c|c|c|}
    \hline
    \multicolumn{5}{|c|}{$Test$ set} \\
	\hline
i & $\mu$  &    Field    &    $\gamma(W_{Sonar})$ & $\tau^{\mu}$\\
\hline
1 & 105 & 1.19697e-01 &  2.09029e-03 & -1 \\
\hline
2 & 107 & 7.97467e-02 &  1.87343e-03 & -1 \\
\hline
3 & 108 & 1.19431e-01 &  1.43453e-02 & -1 \\
\hline
4 & 109 & 3.59889e-02 &  2.09760e-03 & -1 \\ 
\hline
5 & 110 & 1.95963e-02 &  6.21865e-04 & -1 \\ 
\hline
6 & 111 & 6.05680e-02 &  3.18180e-04 & -1 \\ 
\hline
7 & 118 & 2.02768e-02 &  8.74281e-03 & -1 \\ 
\hline
8 & 122 & 3.07859e-02 &  2.66651e-02 & -1 \\ 
\hline
9 & 131 & 6.87472e-02 &  7.54780e-03 & -1 \\ 
\hline
10 & 133 & 1.37587e-02 &  5.23483e-03 & -1 \\ 
\hline
11 & 135 & 4.35705e-03 &  3.23781e-04 & -1 \\ 
\hline
12 & 136 & 7.91603e-03 &  1.01329e-02 & -1 \\ 
\hline
13 & 138 & 2.35263e-02 &  1.13658e-02 & -1 \\ 
\hline
14 & 142 & 2.22331e-02 &  7.53167e-03 & -1 \\ 
\hline
15 & 143 & 2.36318e-02 &  6.63512e-03 & -1 \\ 
\hline
16 & 168 & -1.34434e-02 &  8.57956e-03 & 1 \\ 
\hline
17 & 170 & -8.20828e-02 &  1.96977e-03 & 1 \\ 
\hline
18 & 197 & -4.87395e-02 &  7.08260e-05 & 1 \\ 
\hline
19 & 202 & -2.91468e-03 &  1.02479e-02 & 1 \\ 
\hline
20 & 203 & -8.11795e-02 & 4.08223e-02  & 1 \\ 
\hline
\end{tabular}
\end{minipage}
\hfill
\begin{minipage}[t]{.4\linewidth}
\begin{tabular}{|c|c|c|c|c|}
    \hline
    \multicolumn{5}{|c|}{$Train$ set} \\
	\hline
i & $\mu$ &   Field    &   $\gamma(W_{Sonar})$ & $\tau^{\mu}$\\
\hline
1 &  5  & 1.60234e-02  &   1.99901e-03 & -1 \\
\hline
2 &  6  & 2.76646e-02  &   1.98919e-03 & -1 \\
\hline
3 &  9  & 1.63374e-02  &   5.77784e-05 & -1 \\
\hline
4 &  26 & 1.90089e-02  &   3.73223e-05 & -1 \\
\hline
5 &  39 & 5.77398e-02  &   2.28692e-04 & -1\\ 
\hline
6 &  51 & -5.05614e-02 &    3.23683e-02 &   1\\ 
\hline
7 &  53 & -1.35299e-01 &    2.60559e-03 &  1\\ 
\hline
8 &  55 & -4.13985e-02 &    2.24705e-03 &  1\\ 
\hline
9 &  57 & -7.71201e-02 &    2.30675e-03 &  1\\ 
\hline
10 & 58 & -2.70548e-02 &    2.46977e-03 &  1\\ 
\hline
11 & 61 & -3.02880e-02 &    2.57994e-03 &  1\\ 
\hline
12 & 62 & -3.16819e-02 &    1.94132e-02 &  1\\ 
\hline
13 & 64 & -6.23916e-02 &    7.15758e-03 &  1\\ 
\hline
14 & 65 & -4.00952e-02 &    2.49840e-03 &  1\\ 
\hline
15 & 66 & -2.10826e-01 &   2.43180e-03  & 1\\ 
\hline
16 & 72 & -7.44215e-02 &    2.31141e-02 &  1\\ 
\hline
17 & 73 & -2.71180e-02 &    2.67136e-02 &  1\\ 
\hline
18 & 77 & -1.34531e-01 &    2.44142e-03 & 1\\ 
\hline
19 & 82 & -2.26242e-01 &   1.82462e-03  & 1\\ 
\hline
20 & 83 & -5.91598e-02 &   2.07447e-03  & 1\\ 
\hline
21 & 84 & -5.80561e-02 &    4.04605e-04 &  1\\ 
\hline
22 & 97 & -4.57645e-02 &    4.56010e-04  & 1\\ 
\hline
23 & 98 & -6.72313e-02 &    1.06360e-04  & 1\\ 
\hline
24 & 100 & -1.83484e-02 &    3.35972e-03 & 1\\ 
\hline
\end{tabular}
\end{minipage}
\caption{Bad patterns over $Test$ and $Train$ sets}
\label{tab:Stabilities_tt}
\end{table}

\section{Discussion and Conclusion}

In this paper, we have shown that the sonar benchmark is linearly separable. 
Both sets, Train and Test are it but for hyperplanes differents, it generates a certain $\epsilon_{g}$. We have found solutions to the three sets using the perceptron learning rule \cite{Hertz}, and we have found that the generalization error is superior that the error met with Minimerror.
The weight vector for the complete set, $Sonar$, could be used like test for learning algorithms ables to find the best separator hyperplane.

\end{document}